# Removing Stripes, Scratches, and Curtaining with Non-Recoverable Compressed Sensing

Jonathan Schwartz[1], Yi Jiang[2], Yongjie Wang[3], Anthony Aiello[3], Pallab Bhattacharya[3], Hui Yuan[4], Zetian Mi[3], Nabil Bassim[5], Robert Hovden[1]

[1] *Department of Materials Science and Engineering, University of Michigan, Ann Arbor, MI, USA*

[2] *X-ray Science Division, Advanced Photon Source, Argonne National Laboratory, Argonne, IL, USA*

[3] *Department of Electrical Engineering and Computer Science, University of Michigan, Ann Arbor, USA*

[4] *Canadian Centre for Electron Microscopy, McMaster University, Hamilton, ON, Canada*

[5] *Department of Materials Science and Engineering, McMaster University, Hamilton, ON, Canada*

Corresponding author: jtschw@umich.edu

**Abstract**

Highly-directional image artifacts such as ion mill curtaining, mechanical scratches, or image striping from beam instability degrade the interpretability of micrographs. These unwanted, aperiodic features extend the image along a primary direction and occupy a small wedge of information in Fourier space. Deleting this wedge of data replaces stripes, scratches, or curtaining, with more complex streaking and blurring artifacts—known within the tomography community as 'missing wedge' artifacts. Here, we overcome this problem by recovering the missing region using total variation minimization, which leverages image sparsity-based reconstruction techniques—colloquially referred to as compressed sensing—to reliably restore images corrupted by stripe-like features. Our approach removes beam instability, ion mill curtaining, mechanical scratches, or any stripe features and remains robust at low signal-to-noise. The success of this approach is achieved by exploiting compressed sensing's inability to recover directional structures that are highly localized and missing in Fourier Space.

**Introduction**

Streaks, stripes, scratches and curtaining artifacts commonly degrade image quality in microscopy datasets. This broad class of highly directional artifacts arise from varying conditions during image scanning—or may be artifacts inherent to the specimen, but artificially introduced during sample preparation such as curtaining during ion beam milling or mechanical scratches from polishing techniques. As a result, these artifacts can plague micrographs across any length scale. At the mesoscale, 3D focused ion beam (FIB) tomography is limited by the streaks induced by mill curtaining (Zaefferer et al., 2008). At high-resolution, stripes appear in scanning transmission electron microscopy (STEM) from beam instability and is most noticeable when signal is low relative to the background—common to bright-field (BF) detectors, imaging thick specimens or beam current fluctuation. Even more broadly, stripe artifacts are seen in atomic force microscopes (AFM) (Chen & Pellequer, 2011), light sheet fluorescence microscopy (LSFM) (Liang et al., 2016), and even globally at km length scales in planetary satellite imaging (Rakwatin et al., 2007). When possible, these linear artifacts are best mitigated experimentally, however experimental solutions are often difficult or unavoidable.

Over the years, a few methods have been demonstrated for destriping images outside of electron microscopy. Statistical-based methods developed for multiple-sensor imaging systems in planetary satellites assume the distribution of digital numbers in each sensor should be consistent (i.e. histogram or moment matching) (Rakwatin et al., 2007; Gadallah et al., 2010). However, these matching-based methods are highly limited by the similarity assumption and fail on single-sensor imaging systems. Alternatively, filtering-based methods suppress the presence of stripe noise by constructing a filter on a transformed domain with a Fourier transform (Chen & Pellequer, 2011; Jinsong Chen et al., 2003) or wavelet analysis (Torres & Infante, 2001; Münch et al., 2009). Unfortunately, filtering methods risk removing or suppressing useful structural information falling within the filter. The third approach treats the destriping issue as an ill-posed inverse problem. Prior knowledge is used to regularize an optimization problem (Bouali & Ladjal, 2011) and separate the unidirectional stripes from the image (Liu et al., 2013; Fitschen et al., 2017; Schankula et al., 2018). A similar class of research, known as compressed sensing (CS), has become highly successful toward solving inverse problems with incomplete data by finding maximally sparse solutions — but has yet been applied to remove scratch and stripe artifacts.

In this paper, we describe a compressed sensing inspired approach that can remove highly-directional artifacts and demonstrate applications for ion mill curtaining, mechanical scratches, and beam instability. Here, a wedge of information containing the stripe artifacts is removed in Fourier space and the specimen's information is recovered using total variation (TV) minimization, which maximizes sparsity of the image's gradient magnitude and preserves sharp edges. A data constraint is imposed to produce a stripe free image with near identical appearance in signal-to-noise. The algorithm effectively removes striping when the missing wedge encompasses all stripe artifacts (typically 5º ~ 10º) and is relatively insensitive to noise.

**Background**

The simplest way to remove streaks and stripes is to delete corresponding planes (or wedges) of information in Fourier space, however in turn, this also degrades the image. To illustrate, Figure 1 shows a backscatter electron (BSE) image of a biomineral surface scratched during mechanical polishing. The aperiodic scratches that extend a vertical direction (Fig. 1a) are confined in an angular range (i.e. ~5º) in Fourier space. Removing an information wedge (Figure 1b) can better estimate the true object. However, deleting information in Fourier Space introduces smearing, elongation, and blurring (highlighted in the yellow circles). These missing wedge artifacts are well known to the electron tomography community where larger wedge sizes exacerbate smearing and elongation artifacts (Midgley & Weyland, 2003).

Recently, CS inspired approaches have been applied to tomography as a tool for recovering information in the missing wedge. It is possible to reconstruct models with high data-fidelity from sparse projections under the CS framework. TV minimization is widely used in image restoration because of its ability to preserve edges (Qi et al., 2015). The optimization problem for this algorithm can be written as $min \, \|\nabla \hat{x}\|_1 \, such \, that \, \Phi \hat{x} = b$ where $\hat{x}$ and $b$ represent the reconstructed image and measured data, $\Phi$ is the measurement matrix and $\nabla$ transforms the image to the gradient-magnitude (sparse) domain.

Compressed sensing has demonstrated exact signal recovery is possible from a minimal number of measurements ($b$) by assuming maximal sparsity (Donoho, 2006; Candes et al., 2008). CS solves inverse problems ($\Phi \hat{x} = b$) by seeking the sparsest representation of the original object ($\hat{x}$) via $\ell_1 -$ norm optimization (Natarajan, 1995; Thomas et al., 2013). The theory of CS requires two assumptions to be true: (1) the object must have a sparse representation in a known transform domain (i.e. to be compressible), (2) there must be incoherence (i.e. contain a high level of dissimilarity) between the sensing and sparse basis (Lustig et al., 2007; Candes & Wakin, 2008). Conveniently, experimental data is often sparse in certain domains, as reflected by the compressibility of real-world information (Duarte & Eldar, 2011). Incoherence expresses the level of dissimilarity between the sensing basis and the sparse basis (Leary et al., 2013) and depends on both how information is sampled (or missing) and the structure of the information (i.e. specimen).

One popular sensing basis used in electron microscopy is the Fourier basis. Each plane in Fourier space represents a projection of lines or stripes in real-space (Bracewell, 1956). Thus, extended stripe-like structures in an image are confined to a plane in Fourier space. This is illustrated in the vertical scratches of Figure 1a, which exists in a horizontal plane of the FFT. If there is a small angular range to the stripes, the planes will broaden out to a wedge. Unlike non-directional image features that are typically spread-out in Fourier space, all knowledge of stripes becomes lost by removing a well-chosen plane (or wedge) of information. Thus, without sufficient sampling of particular Fourier planes, linear types of features cannot be retrieved by the standard CS sampling strategy.

In this work we demonstrate that TV minimization advantageously fails to restore stripes and scratches confined within a missing wedge (low incoherence) – but recovers structural information that is relatively spread in Fourier space (high incoherence). This provides the basis to our destriping approach.

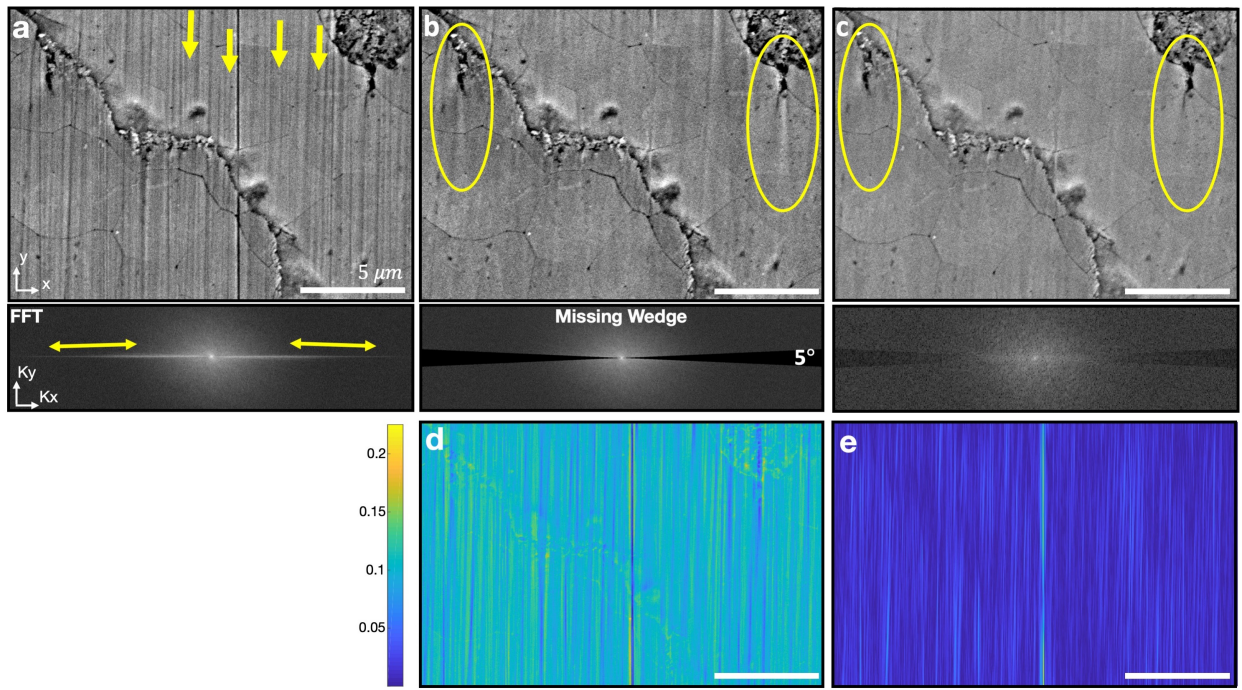

**Fig. 1**. **Reconstruction of a scratched pearl surface**. **a,** The original 10 keV BSE-SEM image of a scratched biomaterial with its FFT below. **b,** The output from deleting a wedge of information in Fourier Space. **c,** The TV minimization reconstruction and its Fourier Space image. **d,** The Residual between images (b) and (a). **e,** The residuals between images (c) and (a).

**Results**

Here we destripe images, by first removing the information in Fourier space containing unwanted artifacts, thus creating a 'missing' wedge. This 'missing' wedge of information is then recovered by minimizing the image's TV using a gradient descent approach. Simply deleting information within the 'missing' wedge in Fourier Space (Figure 1b.) provides a poor estimate of the clean image and creates elongation and blurring artifacts perpendicular to the 'missing' wedge (Fig. 1b). The residuals reveal the quality of the decomposition by calculating the absolute difference between the output and the original image. The residuals from a wedge Fourier filter, for Figure 1b, shows the removal of both unwanted stripes and useful structural information (Fig. 1d), a typical problem associated with filtering. However, recovering information in the 'missing' wedge with TV minimization produces an image without these artifacts (Fig. 1c). Figure 1e shows TV minimization only removed features that strictly pertain to the scratches. The TV minimization algorithm, pseudocode, and parameters is described by Sidky et al. (Sidky et al., 2006) with source code provided in Supplemental Materials. The gradient descent incorporates a convergence parameter ($a$) and requires many iterations to converge. Supplemental Figures S1 and S2 investigate the optimal conditions for convergence and demonstrate that both experimental and simulated data normally converges after ~150 iterations when $a \approx 0.1$.

Other stripe artifacts, such as beam instability common to BF-STEM images (Fig. 2a.) or curtaining in FIB micrographs (Fig 2d.) can also be removed with our approach. Figure 2a. shows an atomic resolution image of an $Al_xGa_{1-x}N$ quantum well suffering from unidirectional horizontal stripes caused by current fluctuations in the beam. The stripes prevent assessment of interface sharpness of the bright Al-rich layer. Similarly, waterfall/curtaining effects (Fig. 2d.) — typically arise in FIB-tomography due to irregular milling rates caused by specimen inhomogeneity (Holzer et al., 2007). The presence of curtaining artifacts reduces interpretability for segmentation or object recognition (Bender et al., 2011). Figures 2b-c and 2e-f demonstrate that our approach can separate the corrupted image into the clean and stripped components. For beam instability, the stripes are unidirectional (i.e. perfectly horizontal) and in this case, the missing wedge is along the x-axis only (no angular spread). In this limiting case, our approach converges to a one-dimensional problem with some similarity to that proposed by (Bouali & Ladjal, 2011; Liu et al., 2013) for Earth satellite and FIB data.

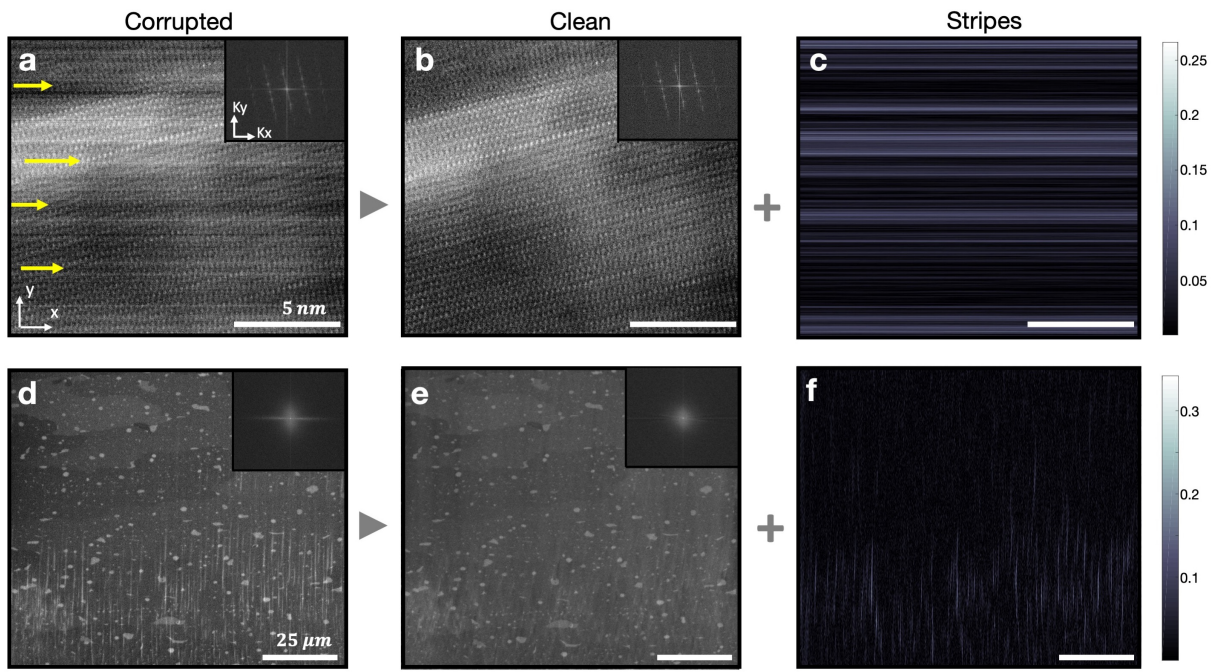

**Fig. 2. Reconstructions of BF-TEM and FIB images. a,** The original contrast reversal 300 keV BF-TEM image of an AlGaN Quantum Well with horizontal intensity fluctuations and its FFT on the top right-hand corner. **b,** The reconstruction with the TV- minimization algorithm and its FFT. **c,** The residuals between images (a) and (b). **d,** The original 30 keV secondary electron (SE)-SEM image of an aluminum sample with curtaining collected during a FIB tomography experiment. **e,** The TV minimization reconstruction and its FFT. **f,** The residuals between images (d) and (e). Contrast was reversed in (a) and (b) for clarity.

An optimal wedge size must be just large enough to remove all the stripe artifacts. Figure 3 shows the performance of removing scratches for 4º, 8º, and 15º missing wedges. If too small of an angle is chosen, then scratch features remain (Fig. 3b1) before and after TV minimization recovery (Fig. 3b2). Thus, a sufficiently large missing wedge should be used to ensure that all information relating to the stripe, scratch, or curtaining has been removed in Fourier space (Fig 3c) and thus cannot be recovered with TV minimization. Increasing the missing wedge of information exacerbates image degradation before TV minimization recovery (Fig. 3b1-d1) but less noticeably after reconstruction. However, one must be cautious not to make the wedge unnecessarily large, as specimen features elongated along the direction of the scratches are susceptible to alteration. The vertical domain boundary highlighted by red circles in Figure 3 is preserved using an optimal angle of ~8º but becomes blurred for much larger angles ~15º.

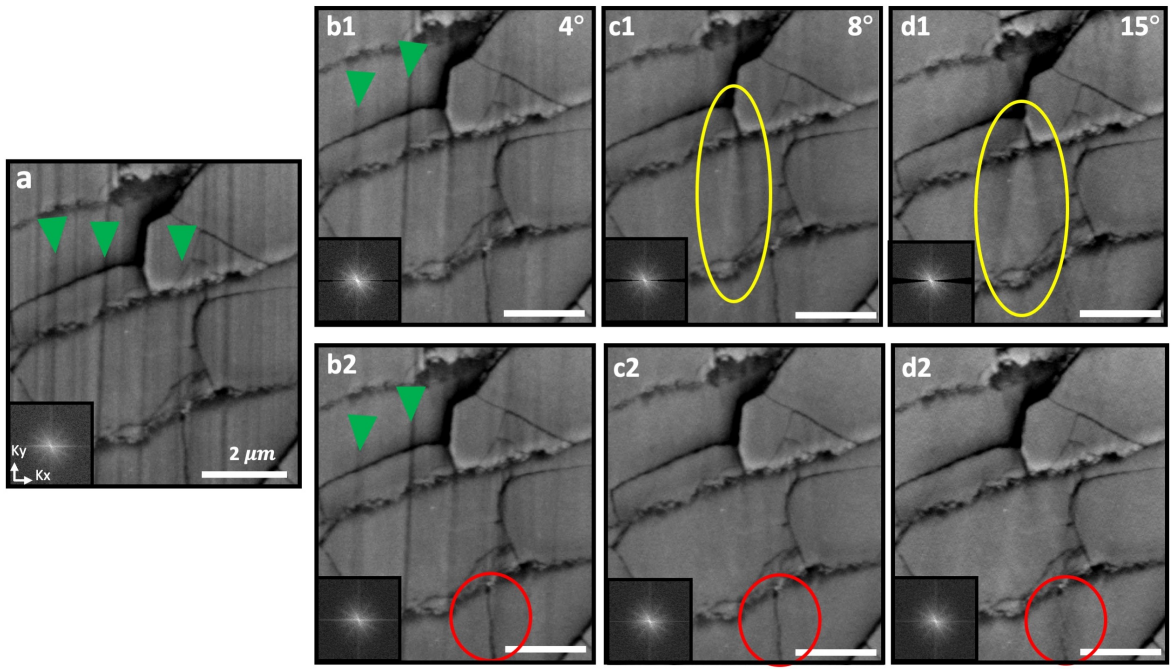

**Fig. 3. Reconstruction of a BSE-SEM micrograph under various horizontal (0º) 'Missing' Wedge widths. a,** The original 10 keV SEM image of a pearl sample with vertical scratches highlighted by green arrows. **b1-d1,** The first row shows the images with a 'missing' wedge of information as the blurring artifacts are prominent in the yellow circles. **b2-d2,** The second row shows the TV minimization reconstructions with the removal of the blurring artifacts and loss of features for large wedges highlighted in the red circles. FFT insets shown lower left.

Our approach is robust even down to low signal to noise ratios (SNR) and has the capability to preserve fine features from the original image. The algorithm consistently produced similar results to the ideal image in Fig. 3c2, after random Gaussian white noise was added to Fig. 4a. The standard deviation ($\sigma$) of gaussian variance of the noise is related to SNR by $\sigma = \frac{\mu}{SNR}$ where $\mu$ is the image's mean. Even though low signal-to-noise (Fig. 4d1) makes the scratches less visible, the algorithm continues to recover the object. Noise preservation is expected from the strict data constraint ($\Phi\hat{x} = b$) that preserves all information in Fourier Space outside of the missing wedge. The goal of this algorithm is not to filter or reduce noise, but to reproduce the original image free from scratches and stripes. Softening the data constraint has been used in electron tomography to smooth tomograms and reduce noise (Jiang et al., 2018). However, the reconstructed image can be over smoothed by TV minimization if the data constraint is too relaxed. Typically, in electron tomography experiments, missing wedges ≳ 25° are recovered with CS. However, here we are showing that highly linear features may not be properly recovered. Because a hard data constraint only allows noise reduction within the small region of the missing wedge (no more than 6% in Fourier Space), the noise structure looks nearly identical to the test images as shown by the reconstructions in Fig. 4.

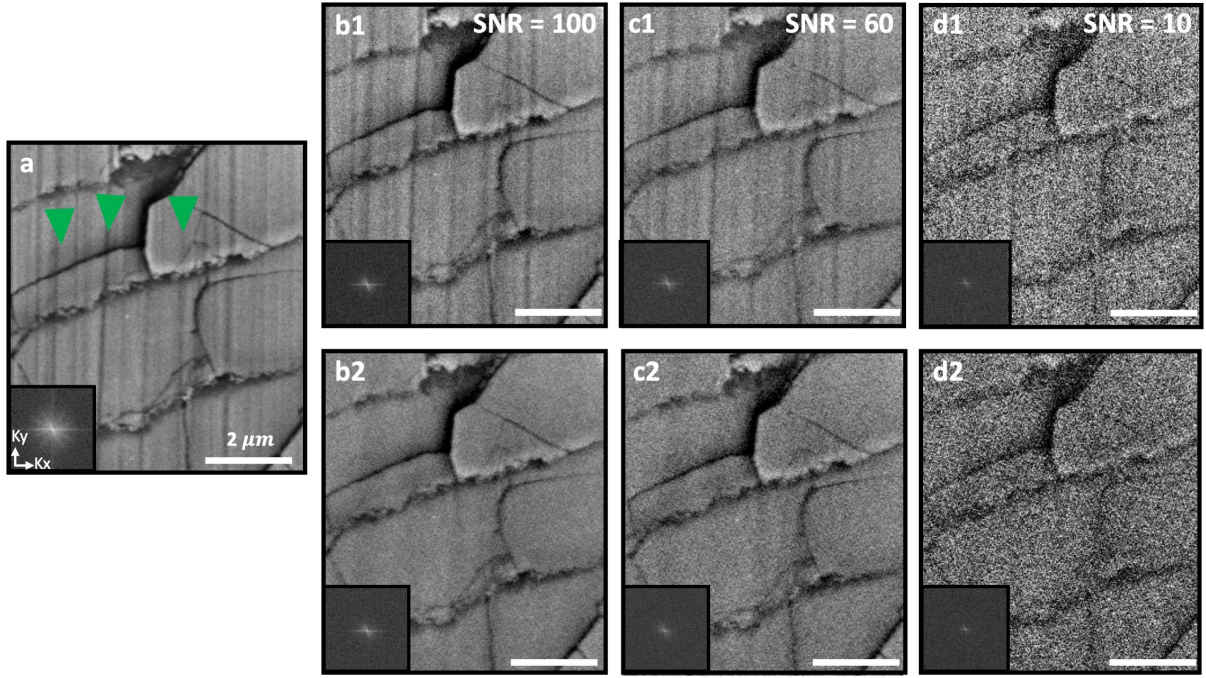

**Fig. 4**. **TV minimization reconstructions for decreasing SNR**. **b1-d1,** The first row shows the images after Gaussian additive white noise is implemented and **b2-d2,** the second row shows the TV minimization reconstructions.

To further understand the relationship between SNR and the missing wedge, a quantitative study was performed on a DF-S/TEM micrograph of InGaN nanowires (Fig. 5b). Figure 5a shows a plot of the reconstruction's root mean square error (RMSE) normalized by the RMSE of the images before reconstruction (i.e. with missing wedge). Pixels with values below 1 indicate improvement from the reconstruction. The typical blurring and elongation artifacts are highlighted by the red circles in Fig. 5c. Similarly to the previous figure, the test image was exposed to random Gaussian white noise to verify the algorithm's performance across multiple SNRs. Our approach best reconstructs the test object for small missing wedges (≲ 12º) and SNR values above 10. The reconstructions are visualized for two wedge sizes and SNR values (Fig.5d-e) to verify the test image (Fig. 5b) is accurately reconstructed while preserving its noise.

Figure S5 and S6 further demonstrates that optimal reconstructions occur across small wedge sizes and down to low SNR.

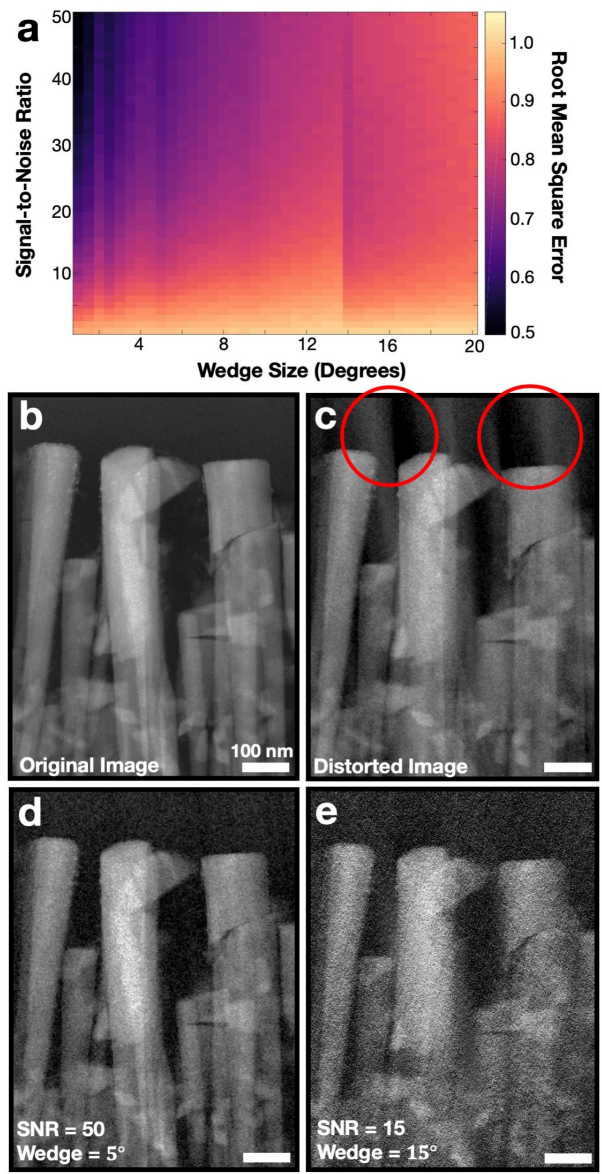

**Fig 5: Quantitative Study of SNR and Wedge Size. a,** A plot of the RMSE normalized by the error from various missing wedge sizes. Values below 1 indicate the reconstruction outperforms loss of information. Wedges below 8º, consistently achieves satisfactory performance at all SNR values above 10. **b,** DF-S/TEM micrograph (at 300 keV) of MBE grown InGaN nanowires with platinum nanoparticles coated on the surface. **c,** A distorted image with a horizontal 5º 'missing' wedge. **d-e,** The outputs for horizontal missing wedges at SNR values of 50 and 15 and missing wedge sizes of 5º and 15º, respectively. Full field of view with the FFTs shown in Supp. Fig. S4.

**Discussion & Conclusion**

In this study, we demonstrate a compressed sensing-based approach to remove highly directional artifacts that commonly occur from ion mill curtaining, mechanical scratches, and beam instability. These highly directional aperiodic features can be removed without introducing blurring or elongation artifacts by removing wedges of information in Fourier space and recovering with TV minimization. Furthermore, our approach remains robust at low SNR. Overall, the qualitative results demonstrate that our technique achieves successful recovery, especially when small wedges are implemented ($\lesssim 10°$). Within electron microscopy, these artifacts may become more common with the rising popularity of FIB sectional tomography that can contain curtaining or monochromated STEM where lens instability causes current fluctuations and image banding. The destriping technique investigated in this manuscript may also have application to a broader field of imaging techniques—such as atomic force microscopy or Raman spectroscopy.

Moreover, this destriping study provides insight to the recoverability of missing wedges in Fourier Space using compressed sensing. Specifically, we show highly directional features that exist within the missing wedge are not recoverable using TV minimization. This has implications for electron tomography, where incomplete experimental measurement commonly results in a missing wedge. Thus, we expect the recovery of highly extended, unidirectional features—such as interfaces—can become difficult for tomography. Acquiring projections perpendicular to interfaces or linear features should overcome this limitation.

**Acknowledgements**

The authors acknowledge support from the Michigan center for Materials Characterization (MC)$^2$ and from DOE Office of Science (DE-SC0011385). Simulations made use of the Advanced

Research Computing Technology Services' shared high performance computing at the University of Michigan.